\documentclass[runningheads]{llncs}
\usepackage[T1]{fontenc}
\usepackage{graphicx}
\usepackage{siunitx}
\usepackage{comment}
\usepackage{amsmath,amssymb}
\usepackage{mathtools}
\usepackage{multirow}
\usepackage{multicol}
\usepackage{amsfonts}
\usepackage{float}
\usepackage{caption}
\usepackage{subcaption}
\usepackage{color}
\usepackage{url}
\usepackage{hyperref}
\usepackage{tabularx,booktabs}
\newcolumntype{Y}{>{\centering\arraybackslash}X}
\usepackage{array,multirow}
\usepackage{multicol}
\usepackage{hhline}

\newcommand\blfootnote[1]{%
  \begin{NoHyper}%
  \renewcommand\thefootnote{}\footnote{#1}%
  \addtocounter{footnote}{-1}%
  \end{NoHyper}%
}
\usepackage{color}
\usepackage{amsmath}
\DeclareMathOperator*{\argmax}{argmax}
\DeclareMathOperator*{\argmin}{argmin}

\usepackage{lipsum}

\newif\ifreview
\reviewtrue
\reviewfalse

\ifreview
	\usepackage{lineno}

	\linenumbers
\fi

\begin{document}

\def\SubNumber{31}
\def\GCPRTrack{Main Track}

\title{Decoupling of neural network calibration measures}

\ifreview
\else
	\author{Dominik Werner Wolf\;\inst{1,2,*} \and
        Prasannavenkatesh Balaji\;\inst{1,3,*} \and Alexander Braun\;\inst{4} \and Markus Ulrich\;\inst{2}}
	\authorrunning{D.~W.~Wolf et al.}
	\institute{GLASS Laboratory, Volkswagen Group, Wolfsburg, Germany
        \email{dominik.werner.wolf@volkswagen.de}\\ \and
        Machine Vision Metrology, Karlsruhe Institute of Technology, Germany \and
        Leibniz University Hannover, Germany \and University of Applied Sciences Duesseldorf, Germany}
\fi

\maketitle
\begin{abstract}
A lot of effort is currently invested in safeguarding autonomous driving systems, which heavily rely on deep neural networks for computer vision. We investigate the coupling of different neural network calibration measures with a special focus on the Area Under the Sparsification Error curve~(AUSE) metric. We elaborate on the well-known inconsistency in determining optimal calibration using the Expected Calibration Error~(ECE) and we demonstrate similar issues for the AUSE, the Uncertainty Calibration Score~(UCS), as well as the Uncertainty Calibration Error~(UCE). We conclude that the current methodologies leave a degree of freedom, which prevents a unique model calibration for the homologation of safety-critical functionalities. Furthermore, we propose the AUSE as an indirect measure for the residual uncertainty, which is irreducible for a fixed network architecture and is driven by the stochasticity in the underlying data generation process~(aleatoric contribution) as well as the limitation in the hypothesis space~(epistemic contribution).

\end{abstract}

\section{Introduction\protect\blfootnote{* Authors contributed equally.}}
Calibrated uncertainties are an essential requirement for sensor fusion. If multiple, independent sensor input signals are processed by neural networks, then the fusion of the feature attributes or the fusion of the ensemble of predictions needs to include the corresponding prediction or embedding uncertainties in order to obtain the most reliable output estimate. Uncertainties in neural networks are generally decomposed into aleatoric and epistemic uncertainty~\cite{Uncertainty_decomposition}. The epistemic uncertainty contribution comprises the model uncertainty and the approximation uncertainty and is generally reducible by considering more sophisticated model architectures or utilizing more training data. On the other hand, the aleatoric uncertainty contribution arises if a non-deterministic prediction problem is studied. Consequently, the variance of the conditional probability $\mathcal{P}\left( y|x \right)$ is non-zero, which results in an irreducible uncertainty contribution to the total prediction uncertainty given by the Shannon entropy.

There are several measures and methods to estimate those uncertainty contributions. Distributional methods like Dropout~\cite{Dropout}, Deep ensembles~\cite{Deep_Ensembles} or the Laplace approximation~\cite{Laplace_approximation} pave the way for the quantitative evaluation of the aleatoric and the epistemic uncertainty contribution. Unfortunately, recent studies~\cite{Uncertainty_Disentanglement} showed evidence on the violation of the disentanglement between the deduced uncertainty contributions, which indicates an inherent inconsistency in the above-mentioned methods.

Nevertheless, the total predictive uncertainty is quantifiable by the Shannon entropy~\cite{Uncertainty_decomposition}. From a metrological point of view, it is expected, that the uncertainty measure coincides with the observed error rate in the network's predictions if the model is properly calibrated. The calibration quality can be evaluated by different metrics, e.g. by the Uncertainty Calibration Error~(UCE)~\cite{mECE_mUCE} or the Area Under the Sparsification Error curve~(AUSE)~\cite{dreissig2023calibration} utilizing the Shannon entropy for sorting. The bottleneck of this approach can be illustrated by a thought experiment. Consider a classification problem with three different classes. If the softmax output assigns $95\%$ of the probability mass to the most likely outcome then, according to the Shannon entropy, the total predictive uncertainty lies in the range of $\left[18.1\%,\;21.2\%\right]$. The degree of freedom in this setup lies in the allocation of the probability mass over the two remaining classes. If the trainable neural network parameters are determined by minimizing the negative log-likelihood, which is equivalent to minimizing the cross-entropy loss during training, then only the softmax output regarding the groundtruth label is considered. The distribution of the remaining probability mass over the wrong classes does not influence the loss at all. This gives rise to the idea of exclusively utilizing the maximum softmax output as a measure of prediction confidence or uncertainty, as defined by the variation ratio~\cite{maag2020time}. By doing so, the dependency on the probability mass distribution over the remaining classes is lifted and the degree of freedom is annihilated. In order to quantify the calibration quality of those point-wise predictive uncertainty estimators, different metrics, such as the Expected Calibration Error~(ECE)~\cite{mECE_mUCE} or the AUSE utilizing the variation ratio for sorting, can be assessed. This paper takes up on the conceptual differences between point-wise predictive uncertainty calibration estimators~(ECE,~AUSE$_{V}$) and entropy-based calibration measures~(UCE,~AUSE$_{S}$) to investigate the coupling mechanism among different calibration metrics.

The calibration of the uncertainty estimates is of paramount importance since neural networks tend to be highly overconfident~\cite{guo2017calibration}. Within the zoo of calibration techniques, Temperature scaling~\cite{guo2017calibration} is one of the most prominent post-hoc techniques. The optimal temperature is determined by minimizing the negative log-likelihood. If all calibration measures were coupled, then one would expect that the temperature that minimizes the logits-based negative log-likelihood~\cite{guo2017calibration} simultaneously imposes a global minimum in the ECE curve over temperature. This should also hold true for alternative measures attempting to estimate the calibration quality.
Our main contributions are:
\begin{itemize}
     \item We investigate and corroborate with previous research on the coupling of the aforementioned metrics (and highlight the issues some of them entail) by conducting experiments with a UNET model~\cite{UNET} trained on the A2D2 dataset~\cite{A2D2} for semantic segmentation. Those experiments will lead us to the conclusion that there is no coherent coupling between the calibration metrics in terms of temperature scaling.
     \item We propose the AUSE methodology, which has been used in the work of Dreissig et al.~\cite{dreissig2023calibration} as a calibration measure, as an estimator for the residual uncertainty by utilizing the negative log-likelihood as the uncertainty measure for sorting the predictions.
\end{itemize}
We will provide empirical evidence for our conclusions based on experiments conducted by the utilization of a UNET architecture for semantic segmentation but we hypothesize that the conclusions are generalizable to different classification tasks as well.

With our contributions, we point out a very critical inconsistency between commonly used calibration metrics for classification tasks. This inconsistency aggravates the situation of quantifying calibrated uncertainties with universal validity. Hence, the fusion process underlies a degree of freedom, which prevents the homologation of safety-critical applications in autonomous driving.

\section{Theory and related work}
\subsection{Calibration approaches}
\label{ssec:calibration_approaches}
Since modern neural networks tend to produce overconfident predictions as outputs~\cite{guo2017calibration}, quantifying the calibration quality (or conversely the calibration error) is essential in perception tasks pertaining to safety-critical domains like autonomous driving~\cite{Wolf_ICCV2023} or medical imaging~\cite{bishop2006pattern}.
A neural network's output prediction vector $\mathbf{\hat{p}}(\boldsymbol{y} | \boldsymbol{w}, \boldsymbol{x})$, for a multi-class setting with~$\boldsymbol{y} \in \mathcal{Y}_K$ classes and input instance~$\boldsymbol{x}$ as well as trained weights~$\boldsymbol{w}$, is interpreted as a confidence vector when the prediction logits from the penultimate layer $\mathcal{L}(\boldsymbol{w}, \mathbf{x})$ are passed into a sigmoidal function like the softmax $\sigma$.


Post-hoc methods to calibrate a trained model's predictions are broadly classified into non-parametric and parametric calibration methods. Generally, non-parametric methods do not preserve the model accuracy, which is undesirable. Parametric approaches like Platt scaling~\cite{platt1999probabilistic} and its multi-class extension temperature scaling~\cite{guo2017calibration} are popular yet straight-forward approaches that employ a global scalar parameter referred to as the temperature~$T$ to rescale the logits vector produced by the penultimate layer of the model. An advantage of temperature-scaling based methods is that logit-rescaling does not alter the ranking of the prediction vector $\mathbf{\hat{p}}(\boldsymbol{y} | \boldsymbol{w}, \boldsymbol{x})$, thereby preserving the model accuracy. This invariance of the temperature scaling methodology regarding the predicted class manifests itself mathematically as
\begin{equation}
    \argmax\limits_K\;\sigma\left(\mathcal{L}\left(\boldsymbol{w}, \boldsymbol{x}\right)\right) \stackrel{acc}{=} \argmax\limits_K\;\sigma\left(\frac{\mathcal{L}\left(\boldsymbol{w}, \boldsymbol{x}\right)}{T}\right)\;.
\end{equation}
Although recent empirical studies have shown that temperature scaling can be surpassed in performance by certain training-time approaches like label-smoothing~\cite{muller2019does}, these are computationally intensive and also unfeasible to apply to black-box models~\cite{mukhoti2020calibrating}. This supports the continued use of temperature scaling and its successor (in terms of improved expressiveness) parameterized temperature scaling in ongoing research~\cite{tomani2022parameterized}.

Nixon et al.~\cite{nixon2019measuring} expose drawbacks of the ECE in estimating the true calibration error, attributing the binning of skewed data into a fixed calibration range as one of the causes. Also, they empirically observe that the calibration loss surface has its respective minimum for the Negative Log-Likelihood~(NLL) and the ECE at different temperatures, hinting at the decoupling between different metrics that this work targets to extend.

\subsection{Calibration metrics}
\label{ssec:calibration metrics}
The metrics that have been employed in estimating the calibration quality in this work are introduced subsequently. Let us consider a model~$\mathcal{M}$ with the weights' vector denoted by $\boldsymbol{w}$ that produces a probability mass vector $\mathbf{\hat{p}}(\boldsymbol{y} | \boldsymbol{w}, \boldsymbol{x})$ over the classes $\mathcal{Y}_K$. The case of perfectly calibrated output probabilities is given if
\begin{equation}
    \mathbb{P}\left(\hat{y} = \mathrm{argmax} \; \mathbf{\hat{p}}(\boldsymbol{y} | \boldsymbol{w}, \boldsymbol{x}) \mid c = \mathrm{max} \; \mathbf{\hat{p}}(\boldsymbol{y} | \boldsymbol{w}, \boldsymbol{x})\right) = c\;, \quad \forall c \in[0,1]\;.
\label{eq:perfect_calibration_ece}
\end{equation}
Here, $\hat{y} \in \mathcal{Y}_K$ is the most likely predicted class, $\hat{p} \in [0,1]$ is the predicted confidence for that class and $c$ is the model accuracy~\cite{guo2017calibration}. This implies that if the model prediction is made with a confidence of $\mathrm{max} \; \mathbf{\hat{p}}(\boldsymbol{y} | \boldsymbol{w}, \boldsymbol{x}) = 80\%$, then in 80 out of 100 repeated experiments the average accuracy should also amount to~$c = 80\%$. Hence, the model accuracy serves as the expected model confidence.

A complementary approach to quantify the miscalibration in uncertainty was introduced as a direct extension of Equation~(\ref{eq:perfect_calibration_ece}) by Laves et al.~\cite{laves2019well}, where it was formulated that a model with an uncertainty of $S\left(\mathbf{\hat{p}}(\boldsymbol{y} | \boldsymbol{w}, \boldsymbol{x})\right) = 20\%$, associated with its predictions over 100 repeated experiments, would similarly lead to an average error of $q = 20\%$, expressed by
\begin{equation}
    \mathbb{P}\left(\hat{y} \neq \mathbb{E}[\hat{y}] \mid S\left(\mathbf{\hat{p}}(\boldsymbol{y} | \boldsymbol{w}, \boldsymbol{x})\right)=q \right)=q\;, \quad \forall q \in[0,1]\;.
    \label{eq:perfect_calibration_uce}
\end{equation}
Here,~$S$ is the metric used to quantify the predictive uncertainty. Shannon entropy is a commonly used measure for the predictive uncertainty due to certain advantageous properties: non-negativity, reaching a maximum for the case of a uniformly distributed $\mathcal{P}(\boldsymbol{y} | \boldsymbol{x})$ and invariance to permutations in the label space~$\mathcal{Y}$~\cite{cover1999elements}. The Shannon entropy~$\mathcal{H}$ in the discrete case (normalized by $\log(K)$ - which is the entropy for a uniform distribution over $K$ classes) is defined for each prediction instance as
\begin{equation}
    \mathcal{H}\left[\mathbf{\hat{p}}(\boldsymbol{y} | \boldsymbol{x})\right] = -\frac{1}{\log\left(K\right)} \sum_{i=1}^{N} \hat{p}_{i}(y_{i} | \boldsymbol{x}) \cdot \log\left(\hat{p}_{i}(y_{i} | \boldsymbol{x})\right)\;.
    \label{eq:Shannon_entropy}
\end{equation}
In the continuous case, the marginal distribution~$\mathcal{P}(\boldsymbol{y} | \boldsymbol{x})$ is obtained by marginalization. On the contrary, the distribution over the weights~$\mathcal{P}(\boldsymbol{\omega})$ is approximated by considering a subsampled set drawn from an ensemble of independently trained neural networks of the same architecture in the discrete case. Since we do not intend to focus on the approximation uncertainty component of the epistemic uncertainty contribution, we assume that~$\mathcal{P}(\boldsymbol{\omega})$ is given by a Dirac delta distribution~$\delta(\boldsymbol{\omega}-\boldsymbol{\hat{\omega}})$, where~$\boldsymbol{\hat{\omega}}$ indicates the obtained point predictions for the weights of the trained UNET.

\subsubsection{Expected Calibration Error:}
The expected calibration error (ECE), introduced by Naeini et al.~\cite{mECE_mUCE} is the weighted and binned average of the difference between the model accuracy and the softmax likelihood of the prediction. The ECE is defined as
\begin{equation}
    \mathrm{ECE}=\sum_{m=1}^M \frac{\left|B_m\right|}{n}\left|\operatorname{acc}\left(B_m\right)-\operatorname{conf}\left(B_m\right)\right|\;,
\label{eq:ece}
\end{equation}
where the model's predictions from the softmax function~$\sigma$ are binned into~$M$ bins $B_m$ of equal width in the range~$[0,1]$ and the total number of predictions is denoted by $n$. The absolute difference between the bin-wise average accuracy $\operatorname{acc}\left(B_m\right)$ and average confidence $\operatorname{conf}\left(B_m\right)$ weighted by the fraction of predictions present in each bin results in the ECE.

\subsubsection{Uncertainty Calibration Error:}
Analogous to the definition of ECE in Equation~(\ref{eq:ece}) and its interpretation, the expected uncertainty calibration error~(UCE) can be expressed as the weighted and binned average of the difference between the model error (modeled as the complement to the model accuracy) and the total uncertainty in the prediction (usually quantified by the Shannon Entropy). Mathematically, the UCE is given by
\begin{equation}
    \mathrm{UCE}=\sum_{m=1}^M \frac{\left|B_m\right|}{n}\left|\operatorname{err}\left(B_m\right)-\operatorname{unc}\left(B_m\right)\right|\;.
\label{eq:uce}
\end{equation}
Here, $\operatorname{err}\left(B_m\right) \coloneqq 1/\left|B_m\right| \sum_{i \in B_m} \mathbf{1}\left(\hat{y}_i \neq \mathbb{E}[\hat{y}_i]\right)$ determines the binwise error and $\operatorname{unc}\left(B_m\right) \coloneqq 1/\left|B_m\right| \sum_{i \in B_m} \mathcal{H}_{i}\left[\mathbf{\hat{p}}_{i}(\boldsymbol{y}_{i} | \boldsymbol{x}_{i})\right]$ defines the binwise total uncertainty.

\subsubsection{Uncertainty Calibration Score:}
Originally introduced by Wursthorn et al.~\cite{wursthorn2024uncertainty} for estimating uncertainty calibration in a pose estimation problem, the Uncertainty Calibration Score~(UCS) is defined as
\begin{equation}
    \mathrm{UCS} = 1 - \frac{A}{A_{max}}\;,
\label{eq.ucs}
\end{equation}
where $A$ denotes the area enclosed between the reliability curve and the expected diagonal in case of perfect calibration. The worst possible case of calibration corresponds to the maximum of this enclosed area $A_{max}$ which is 0.25~\cite{wursthorn2024uncertainty}.
As we deal with two terminologically similar calibration metrics in this work (ECE and UCE), we propose to differentiate between these and reinterpret the UCS as CCQS (Confidence Calibration Quality Score) and UCQS (Uncertainty Calibration Quality Score) obtained from the confidence-based and uncertainty-based calibration curves respectively.

\subsubsection{Area Under Sparsification Error:}
\label{sssec:ause}
Sparsification refers to the sequential removal of elements from the validation set, sorted in descending order based on a predictive uncertainty measure. A sparsification curve is computed as a function of sparsification. In the context of multi-class classification, if such an uncertainty measure that attempts to estimate the uncertainty works optimally as intended, then the wrongly classified instances (the sum of false positives and false negatives) are targeted first by the measure. As soon as the sparsification process has removed all wrongly classified instances, the target key performance indicator should be maximized. A sparsification curve obtained from such an flawless uncertainty measure is the $oracle$ curve. It is the theoretical upper bound of all sparsification curves.

The sparsification error curve is obtained as the difference between the oracle curve and the sparsification curve utilizing a predictive uncertainty measure. It is common to use the Shannon entropy from Equation~(\ref{eq:Shannon_entropy}) as an uncertainty measure to sort and subsequently sparsify the predictions. Finally, a proper scoring rule like the Brier score~\cite{gneiting2007strictly} can be used to evaluate the remaining predictions in the validation set~\cite{Gustafsson_2020_CVPR_Workshops}. An alternative approach is to evaluate the sparsified predictions directly with the IoU and to interpret the complement to the softmax likelihood as the predictive uncertainty measure, referred to as the Variation Ratio~(V)~\cite{maag2020time} given by
\begin{equation}
    V = 1 - \mathrm{max} \; \mathbf{\hat{p}}(\boldsymbol{y} | \boldsymbol{w}, \boldsymbol{x})\;.
\label{eq:variation ratio}
\end{equation}
Dreissig et al.~\cite{dreissig2023calibration} employ this approach since it eliminates the conundrum that arises out of the difference between the nature of uncertainty quantified by Shannon entropy and the variation ratio - while the entropy considers the entire probability vector, the variation ratio considers only the uncertainty associated with the top-1 class predicted by the model~\cite{dreissig2022calibration}.
Finally, the integral of the sparsification error curve, defined by the difference between the oracle and the sparsification curve, results in a scalar calibration measure denoted as the Area Under the Sparsification Error curve~(AUSE).

\subsection{Uncertainty decomposition}
\label{ssec:uqdecomp}
If a neural network is employed to infer information about the output $\mathbf{y}$ from an input instance $\mathbf{x}$, it is desired to simultaneously quantify the corresponding prediction uncertainty. A standard measure for the predictive uncertainty, also referred to as the total uncertainty, is the Shannon Entropy~(\ref{eq:Shannon_entropy}). From the conditional probability of $\mathcal{P}(\boldsymbol{y} | \boldsymbol{w}, \boldsymbol{x})$ we can retrieve the marginal distribution $\mathcal{P}(\boldsymbol{y} | \boldsymbol{x})$ by integrating over the probability distribution of weights $\mathcal{P}(\boldsymbol{w})$
\begin{equation}
    \mathcal{P}(\boldsymbol{y} | \boldsymbol{x}) = \int \mathcal{P}(\boldsymbol{y} | \boldsymbol{w}, \boldsymbol{x}) \cdot \mathcal{P}(\boldsymbol{w}) \; \mathrm{d}\boldsymbol{w}\;.
\label{eq:marginal_distribution}
\end{equation}
\noindent
The probability density function~$\mathcal{P}(\boldsymbol{w})$ is sometimes referred to as the second-order distribution of the outcome $\boldsymbol{y}$. The total uncertainty quantified by the Shannon entropy can be decomposed into the aleatoric and the epistemic contribution as~\cite{Depeweg2017DecompositionOU,Uncertainty_decomposition}
\begin{align}
\begin{split}
    \mathcal{H}\left[\mathcal{P}(\boldsymbol{y}|\boldsymbol{x})\right] &= \int \mathcal{H}\left[\mathcal{P}(\boldsymbol{y}|\boldsymbol{w}, \boldsymbol{x})\right]  \cdot \mathcal{P}(\boldsymbol{w}) \; \mathrm{d}\boldsymbol{w}\\
    &+ \int \mathcal{D}_{\mathrm{KL}}\left[\mathcal{P}(\boldsymbol{y} | \boldsymbol{w}, \boldsymbol{x}) || \mathcal{P}(\boldsymbol{y} | \boldsymbol{x})\right] \cdot \mathcal{P}(\boldsymbol{w}) \; \mathrm{d}\boldsymbol{w}\;.
\end{split}
\label{eq:uncertainty_decomposition}
\end{align}
The first term denotes the expected conditional entropy and describes the alea-toric uncertainty contribution, which refers to the stochasticity in the underlying data-generating process. The second term is determined by the expected Kullback-Leibler divergence and characterizes the epistemic uncertainty component, which quantifies the information gain w.r.t. the model weights $\boldsymbol{w}$ if the unconditional distribution $\mathcal{P}(\boldsymbol{y} | \boldsymbol{x})$ would have been known a priori. The decomposition in Equation~(\ref{eq:uncertainty_decomposition}) considers the distribution of weights given a model architecture. This implicitly limits the hypothesis space. It could even be possible that the hypothesis space does not comprise the groundtruth functional relationship underlying the data generation process. In this case, there is also a model uncertainty that needs to be considered in addition to the approximation uncertainty described by the distribution over the weights $\mathcal{P}(\boldsymbol{w})$. The Bregman decomposition~\cite{Uncertainty_Disentanglement} considers this contribution as a third term in the total uncertainty decomposition. Since the groundtruth function is not known a priori, the Kullback-Leibler divergence between the groundtruth and the posterior distribution needs to be approximated to quantify the bias. In Section~\ref{section:AUSE_uncertainty}, we will present an alternative strategy to estimate this bias term by utilizing the AUSE methodology.

\section{Evaluation setup}
\label{ssec:arch}
This work uses a customized UNET architecture~\cite{UNET} with five downsampling stages. The network comprises 7,271,110~trainable parameters and 1,178~non-trainable parameters, which track the first two statistical moments of the batch distribution during training mode for the batch normalization and are not subject to the loss backpropagation. Furthermore, the depth of the CNN extends over 48~layers plus the input and output layer. The wideness of the network doubles before every downsample step to conserve information and reaches a maximum of 304~channels at the UNET bottleneck. Additionally, the finesse of the input is given by $604\times960$~pixels and reaches a minimum in the UNET bottleneck of $19\times30$~pixels. The weight tensors of the convolutional layers are regularized by kernel orthonormality regularization~\cite{KOR}, which encourages the network to learn orthonormal features. Moreover, the loss function is given by the balanced cross-entropy and utilizes a focal loss modulation in order to down-weight the loss assigned to well-classified instances. The cross-entropy is balanced by weighting the classes according to their frequency in the A2D2 dataset, similar to Dreissig et al.~\cite{dreissig2023calibration}. Finally, the UNET has been trained for 400~epochs on the A2D2~segmentation dataset~\cite{A2D2} by utilizing two Nvidia RTX A6000 with 48GB. The trained UNET achieves a mIoU of $74.8\%$ on the evaluation set.

The evaluation set consists of 200 randomly sampled images from the A2D2 segmentation validation dataset. The validation dataset was generated by considering a test ratio of~$20\%$ w.r.t. the cardinality of the entire A2D2 dataset and is downsampled by a factor of two to match the image resolution that was used during model training. Although the A2D2 dataset provides annotations for 38 classes in its groundtruth data, the groundtruth was relabeled to match the 19~classes labeling taxonomy of the Cityscapes dataset~\cite{Cordts2016Cityscapes}.

\section{Evaluation results}
\subsection{Impact of temperature scaling}
\begin{figure}[!t]
    \centering
    \begin{minipage}{.49\textwidth}
        \centering
        \includegraphics[width=\linewidth]{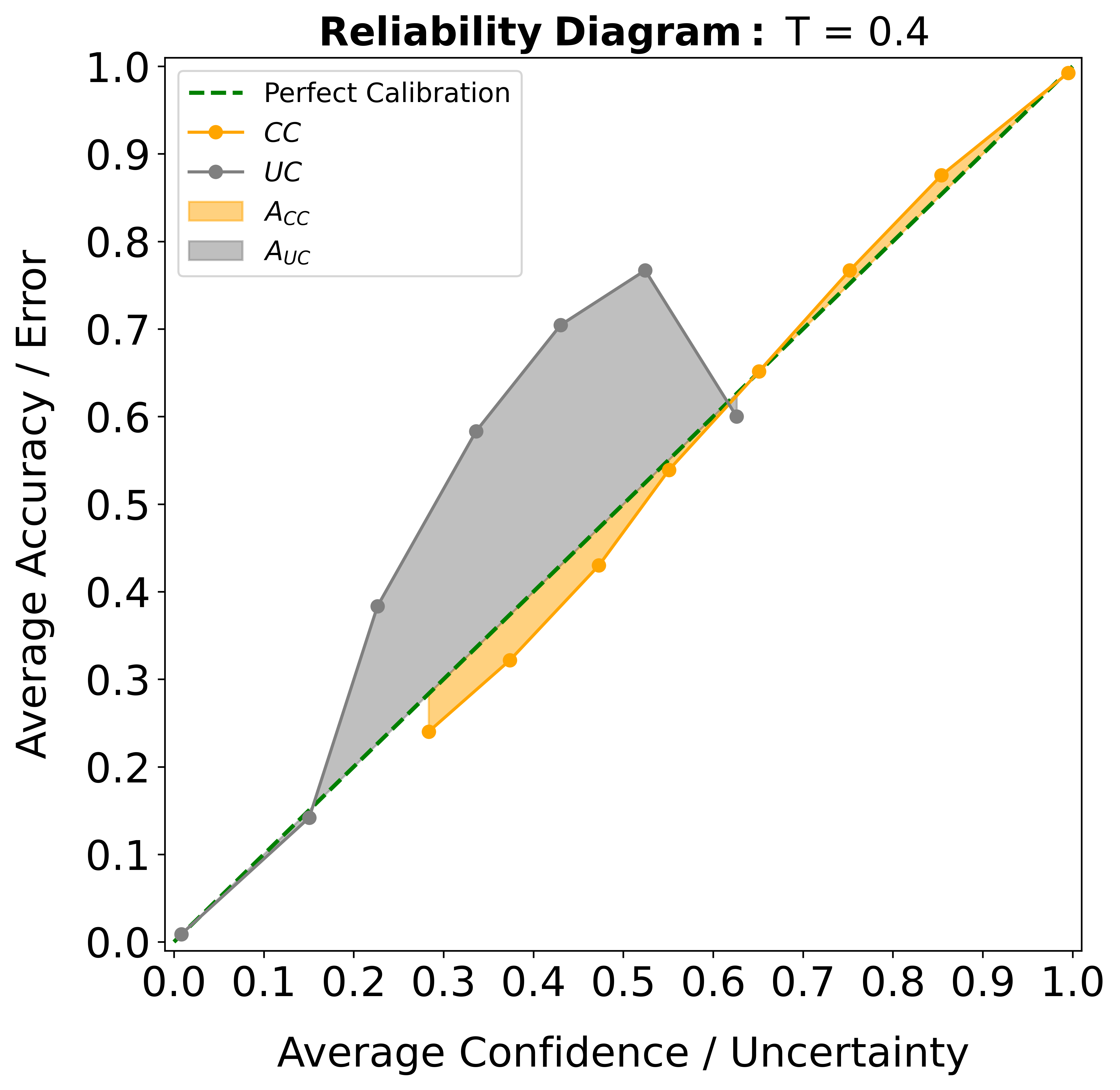}
    \end{minipage}
    \hfill
    \begin{minipage}{0.49\textwidth}
        \centering
        \includegraphics[width=\linewidth]{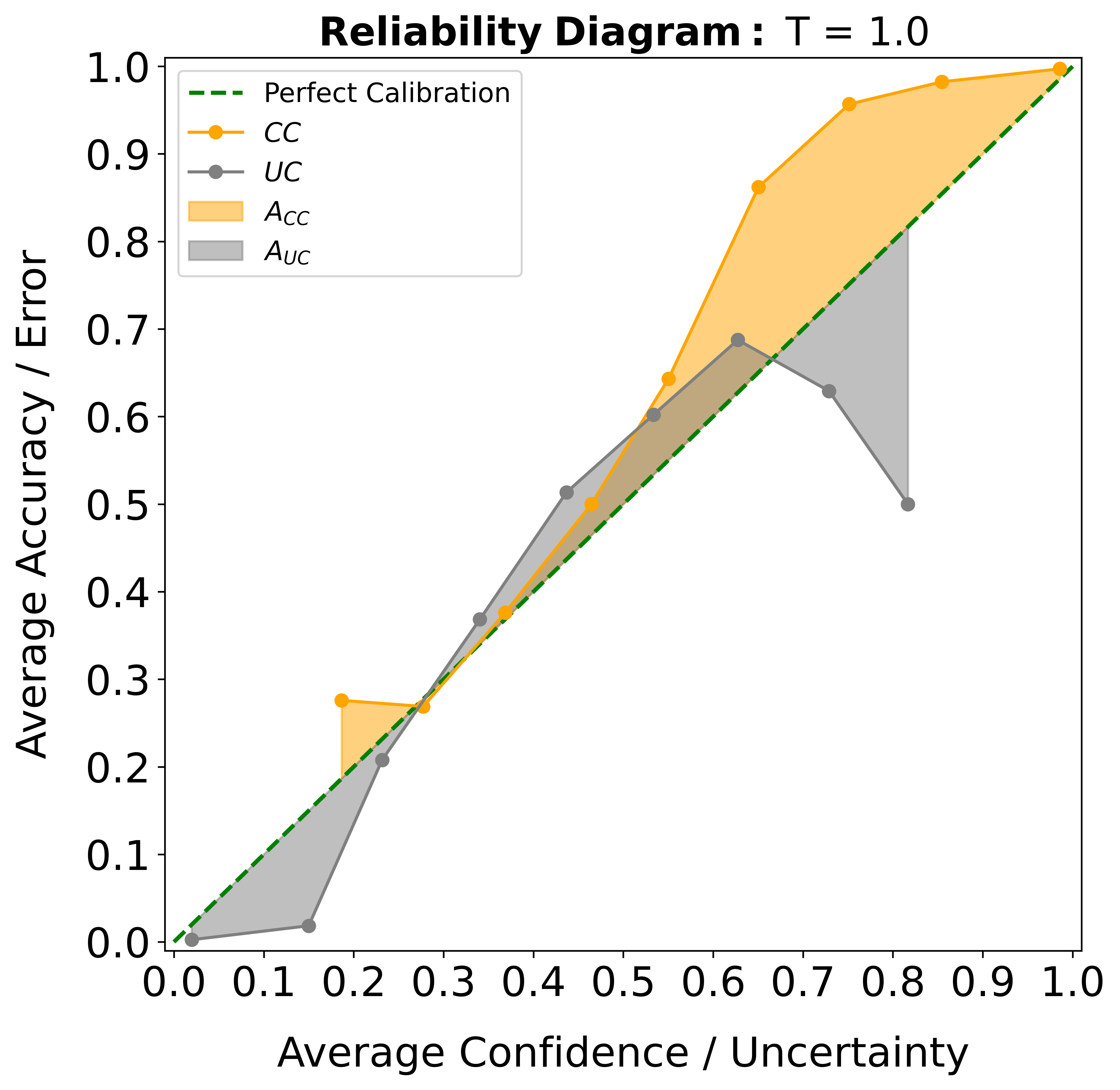}
    \end{minipage}
\caption{The reliability diagrams for the untempered case ($T$ = 1.0, left) and for the optimal temperature (right) are illustrated along with the enclosed areas $A_{CC}$ and $A_{UC}$.}
\label{fig:ts_reliability plots}
\vspace{-0.5cm}
\end{figure}

\begin{figure}[!t]
    \centering
    \begin{minipage}{0.49\textwidth}
        \centering
        \includegraphics[width=\linewidth]{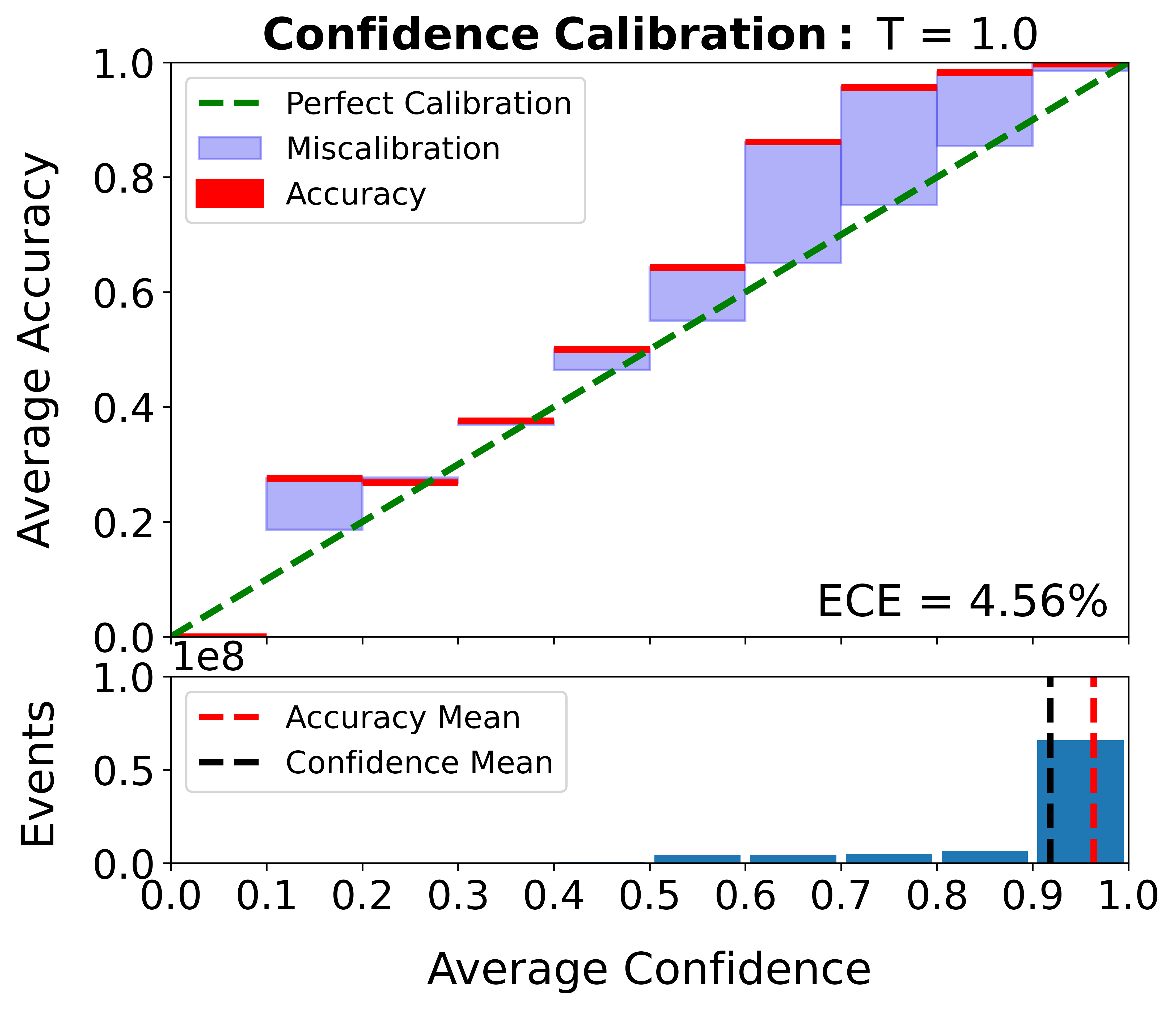}
    \end{minipage}
    \hfill
    \begin{minipage}{0.49\textwidth}
        \centering
        \includegraphics[width=\linewidth]{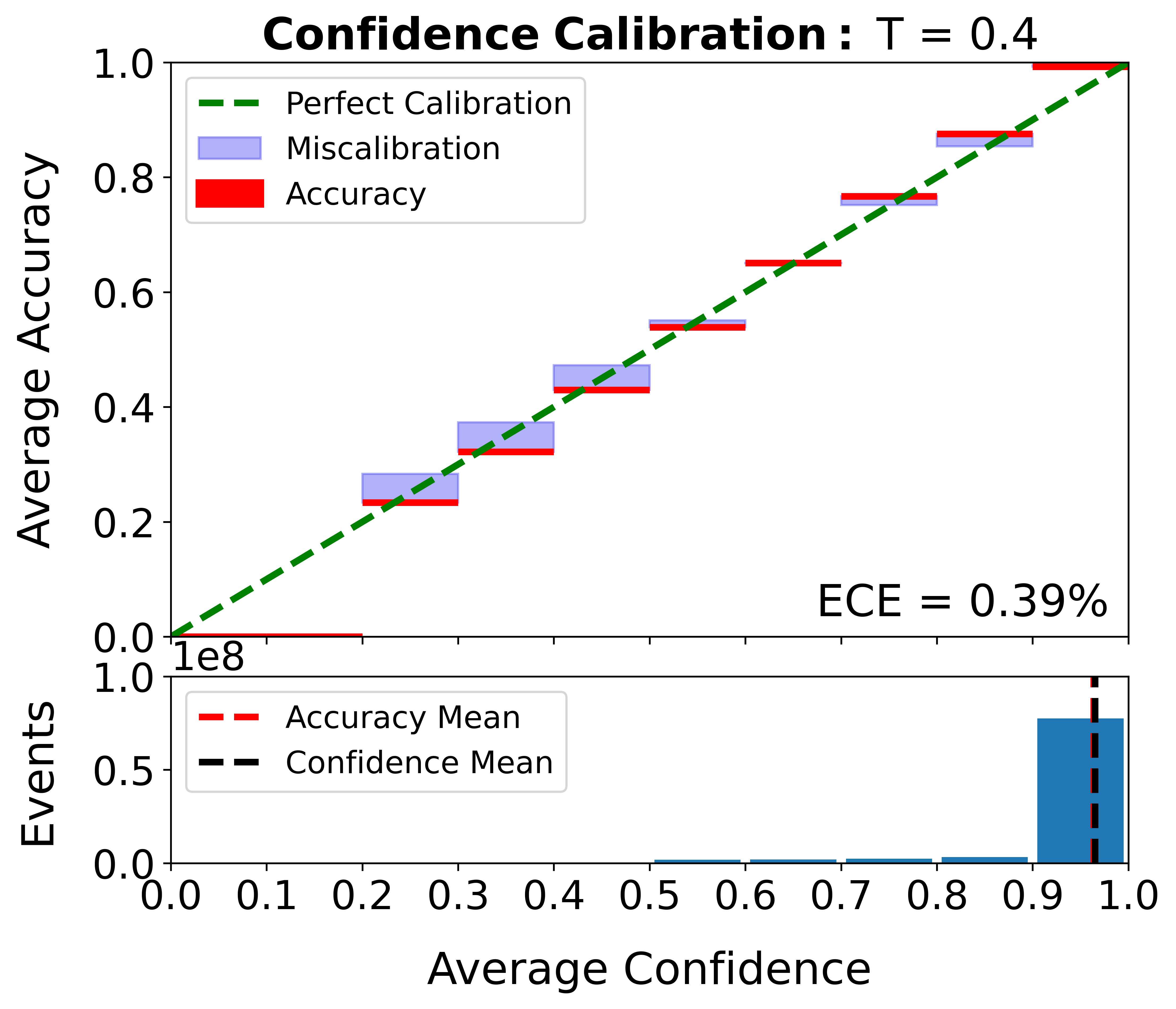}
    \end{minipage}
    \medskip
    \begin{minipage}{0.49\textwidth}
        \centering
        \includegraphics[width=\linewidth]{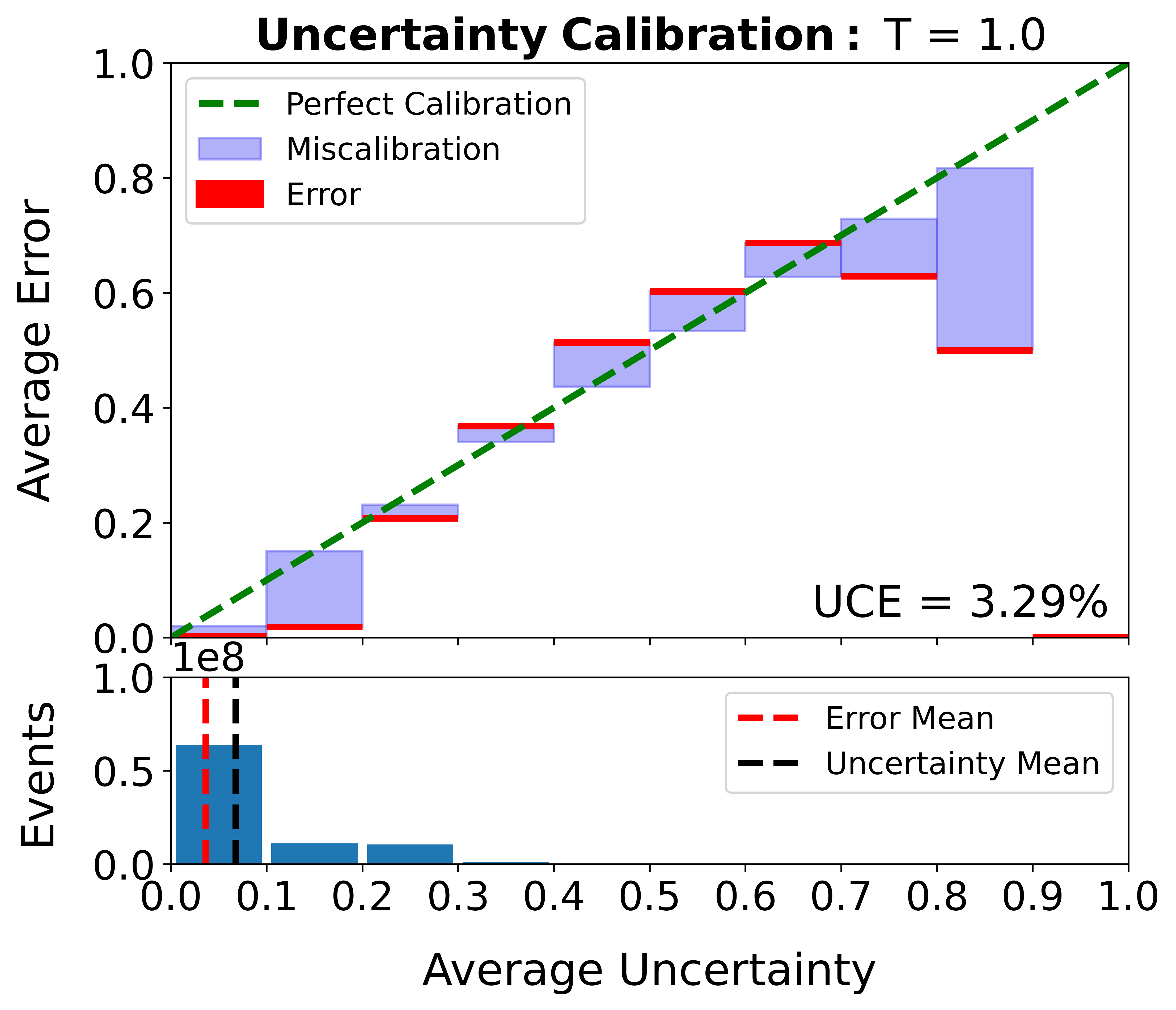}
    \end{minipage}
    \hfill
    \begin{minipage}{0.49\textwidth}
        \centering
        \includegraphics[width=\linewidth]{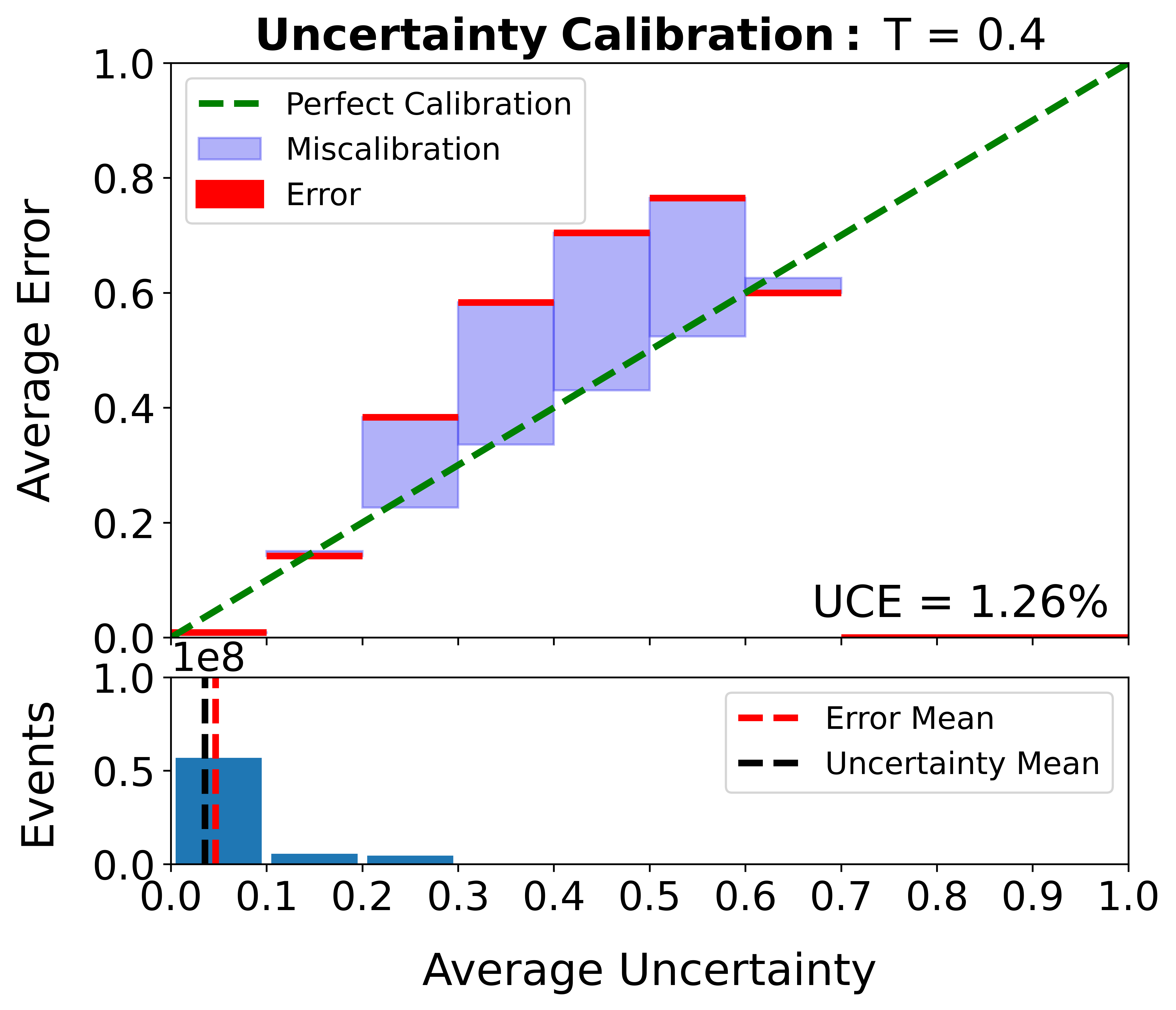}
    \end{minipage}
\vspace{-0.3cm}
\caption{Calibration does not guarantee that the miscalibration in every bin is minimized since the bin-wise distribution of predictions is extremely skewed. This provides evidence on why the CCQS and UCQS for skewed data will not be coherently maximized as the ECE and the UCE are minimized.}
\label{fig:ts_ece_uce}
\end{figure}

The expectation is that all calibration measures discussed in Section~\ref{ssec:calibration metrics} should be coherent in such a way that all measures that intend to estimate the true calibration error in the predictions should approach a minimum at the same temperature. Figure~\ref{fig:ts_reliability plots} illustrates the impact of temperature on the reliability-diagram-based metrics~(ECE, CCQS, UCE and UCQS). The optimal temperature is given by the location of the minimum in the mean ECE over temperature graph, which coincides with the optimal temperature of the mean UCE. It can be noticed, that the enclosed areas $A_{CC}$ and $A_{UC}$, which relate to the CCQS and UCQS, are not simultaneously minimized for the optimal temperature of the ECE/UCE. This is caused by the extreme skewness in the distribution of the predicted average confidences as well as the distribution of the predicted average uncertainties respectively, which is highlighted in Figure~\ref{fig:ts_ece_uce}. The skewness accounts for the degree of asymmetry observed in the frequency distribution and is quantifiable by the third standardized moment. As a consequence, the predictions in the last bin (for the ECE) and the first bin (for the UCE) are dominating the calibration process. In a nutshell, the CCQS or the UCQS as a calibration measure targets the fulfillment of the calibration condition in Equation~(\ref{eq:perfect_calibration_ece}) and in Equation~(\ref{eq:perfect_calibration_uce}) over the entire average confidence/uncertainty range, whereas the ECE and the UCE are implicitly focused to a subdomain due to the skewed frequency distribution. In safety-critical situations, such a focus on highly confident predictions is not favorable because the sparsely observed scenarios with the lowest confidence will be the most critical ones for preventing an accident in autonomous driving.

Figure~\ref{fig:ts_ece_uce} further illustrates the impact of the optimal temperature with the bin-wise gap in calibration indicated along with its frequencies of occurrence.

\subsection{Class-wise analysis}
Four metrics are considered for a class-wise analysis: the ECE (conventionally considering only the top-1 softmax confidence), the UCE (using Shannon entropy) and its two equivalent metrics from the sparsification space, the AUSE calculated using variation ratio as the sorting measure (AUSE$_V$) and the AUSE calculated using Shannon entropy as the sorting measure (AUSE$_S$) respectively. For the computation of the AUSE, the IoU was used as a merit function. The optimal temperature that minimizes these four metrics differs from one class to another. This is attributable to the level of representation of a particular class in the dataset and ultimately to how well each class' features have been learned by the model during training. In Table~\ref{tab:temptable_400epoch}, the optimal temperatures $T_{\mathrm{cal}}$ for the most represented and the least represented classes in the evaluation set are tabulated.

\begin{table}[!h]
    \vspace{0.2cm}
    \centering
    \begin{tabularx}{\textwidth}{ |c| *{8}{Y|} }\cline{1-9}
    
    \multirow{2}{*}{\textbf{Class}} & \multicolumn{4}{c|}{\textbf{Reliability}}      & \multicolumn{4}{c|}{\textbf{Sparsification}} \\\cline{2-9}
                                    & \multicolumn{2}{c|}{$\argmin \limits_{T}$ ECE} & \multicolumn{2}{c|}{$\argmin \limits_{T}$ UCE}  & \multicolumn{2}{c|}{$\argmin \limits_{T}$ AUSE$_V$}  & \multicolumn{2}{c|}{$\argmin \limits_{T}$ AUSE$_S$} \\ \hline

    Nature & 0.4 & \multirow{3}{*}{ $\pm$0.05 } & 0.4 & \multirow{3}{*}{ $\pm$0.05 } & 1.8 & \multirow{3}{*}{ $\pm$0.17 } & 0.9 & \multirow{3}{*}{ $\pm$0.38 } \\ \cline{1-2} \cline{4-4} \cline{6-6} \cline{8-8}
    Sky & 0.4 & & 0.4 & & 1.4 & & 0.9 & \\ \cline{1-2} \cline{4-4} \cline{6-6} \cline{8-8}
    Road & 0.5 & & 0.5 & & 1.7 & & 0.1 & \\ \hhline{|=|=|=|=|=|=|=|=|=|}

    Pedestrian & 0.3 & \multirow{3}{*}{ $\pm$0.31 } & 0.4 & \multirow{3}{*}{ $\pm$0.40 } & 0.9 & \multirow{3}{*}{ $\pm$0.52 } & 0.9 & \multirow{3}{*}{ $\pm$0.17 } \\ \cline{1-2} \cline{4-4} \cline{6-6} \cline{8-8}
    Small Vehicle & 0.2 & & 0.3 & & 1.9 & & 1.0 & \\ \cline{1-2} \cline{4-4} \cline{6-6} \cline{8-8}
    Animal & 0.9 & & 1.2 & & 0.7 & & 0.6 & \\ \hline   
    \end{tabularx}
    \vspace{3mm}
\caption{$T_{\mathrm{cal}}$, which minimizes the respective calibration metric is listed for the three most and least represented classes in the evaluation set~(top three and bottom three respectively). It is noticeable that the standard deviation of the class-wise $T_{\mathrm{cal}}$ is amplified if a subset of less represented classes is considered.}
\label{tab:temptable_400epoch}
\vspace{-0.2cm}
\end{table}

\subsection{Average performance of the calibration measures}
\begin{figure}[!t]
    \centering
    \begin{minipage}{0.45\textwidth}
        \centering
        \includegraphics[width=\linewidth]{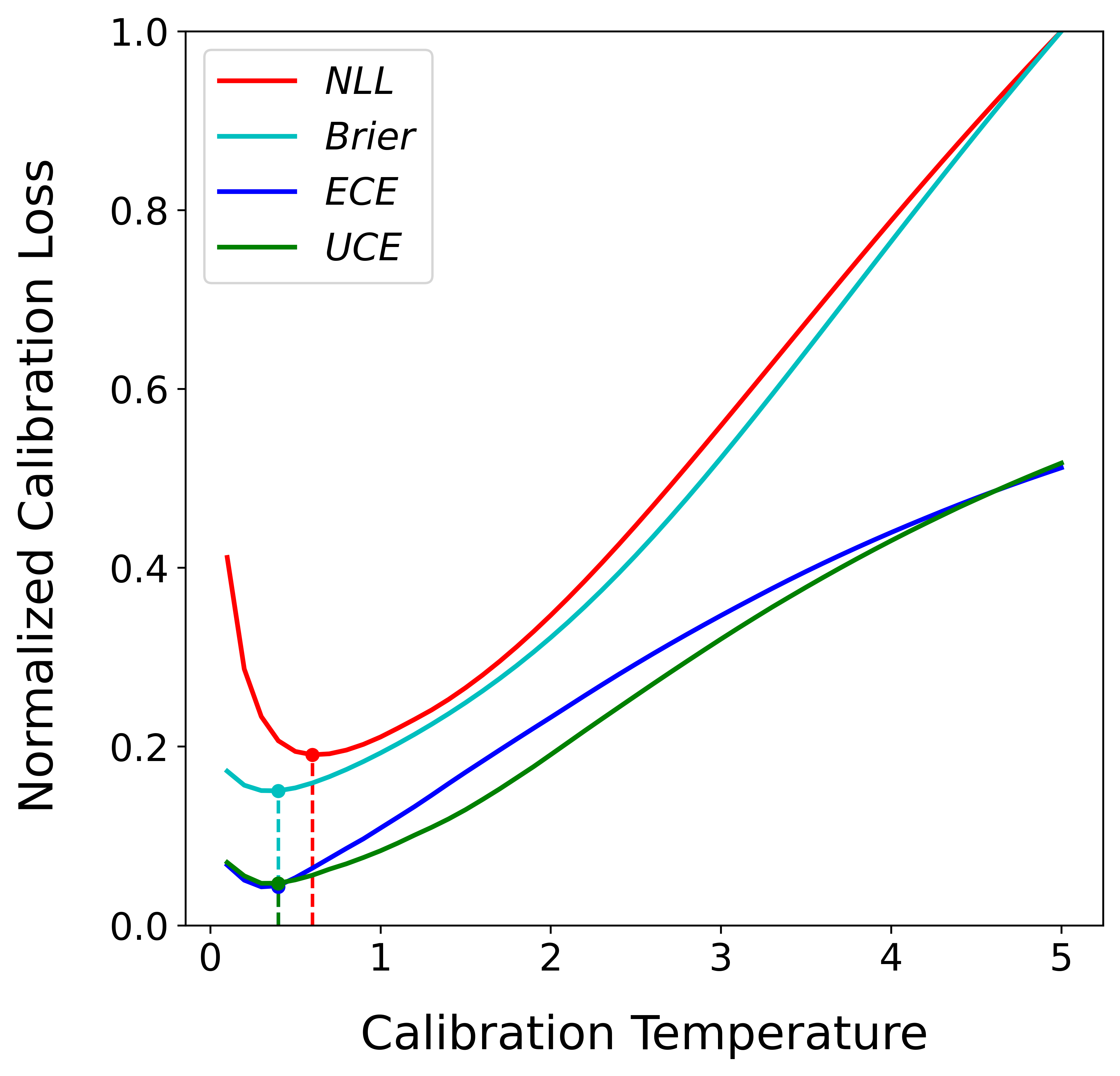}
    \end{minipage}
    \hfill
    \begin{minipage}{0.54\textwidth}
        \centering
        \includegraphics[width=\linewidth]{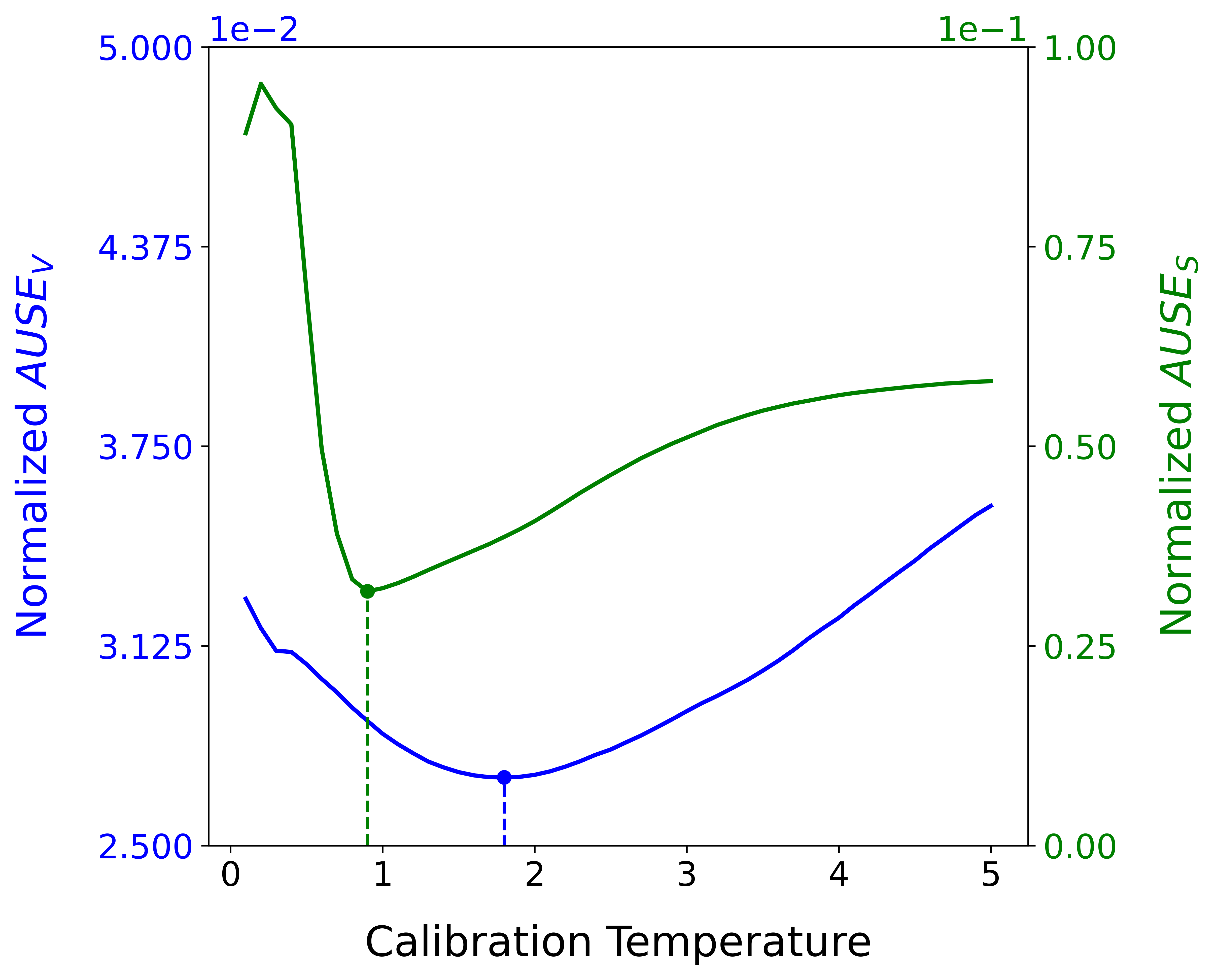}
    \end{minipage}
\caption{Left: The normalized calibration loss surface of NLL, Brier score, the mean of class-wise ECE and the UCE are plotted. The optimal temperature for the Brier score, ECE and UCE coincide at $T$ = 0.4 while the NLL indicates a minimum at $T$ = 0.6. Right: The normalized calibration loss surface of AUSE$_V$ and AUSE$_S$ indicate minima at $T$=1.8 and $T$=0.9 respectively.}
\label{fig:Calibration_minima}
\vspace{-0.4cm}
\end{figure}

To extend our study from the class-wise impact of temperature scaling on calibration and understand the average impact on all classes present in the evaluation set, the mean of each metric (computed class-wise) at all the temperatures in the range of $T \in [0.1, 10]$ was calculated, along with the NLL and the Brier score. For example, the class-wise ECEs are calculated at every temperature and their means are plotted. At this point it has to be emphasized, that there is a difference between calculating the ECE holistically by including all classes at once versus averaging over the class-wise ECE values, as demonstrated by Kull et al.~\cite{kull2019beyond}. Since we study the AUSE by utilizing the IoU as a merit function, a class-wise evaluation is required in order to account for class imbalances, which could ultimately lead to an intersection between the sparsification and the oracle curve that has to be prevented. Consequently, we align to the class-wise analysis and deduce the mean calibration performance by averaging over the class-wise measures. As a side note, we observe in our experiments the same optimal temperature for the holistic ECE/UCE, including all classes at once, and the mean ECE/UCE obtained by averaging over all classes. The average performances of the calibration measures over the temperature are condensed in Figure~\ref{fig:Calibration_minima}.

\subsection{AUSE as an uncertainty estimator}
\label{section:AUSE_uncertainty}
\begin{figure}[!t]
    \centering
    \begin{minipage}{.475\textwidth}
        \centering
        \includegraphics[width=\linewidth]{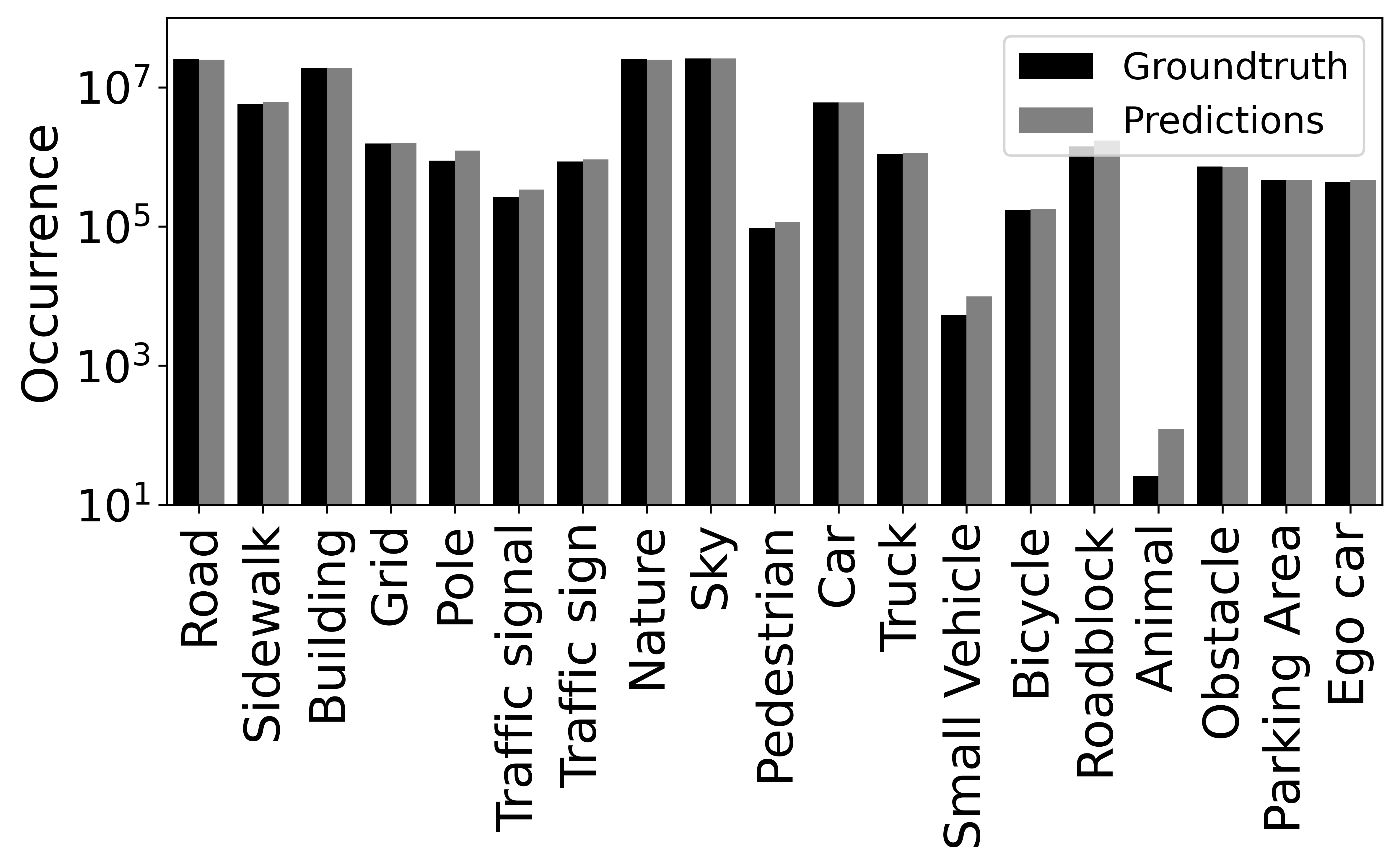}
    \end{minipage}
    \hspace{0.3cm}
    \begin{minipage}{0.4\textwidth}
        \centering
        \includegraphics[width=\linewidth]{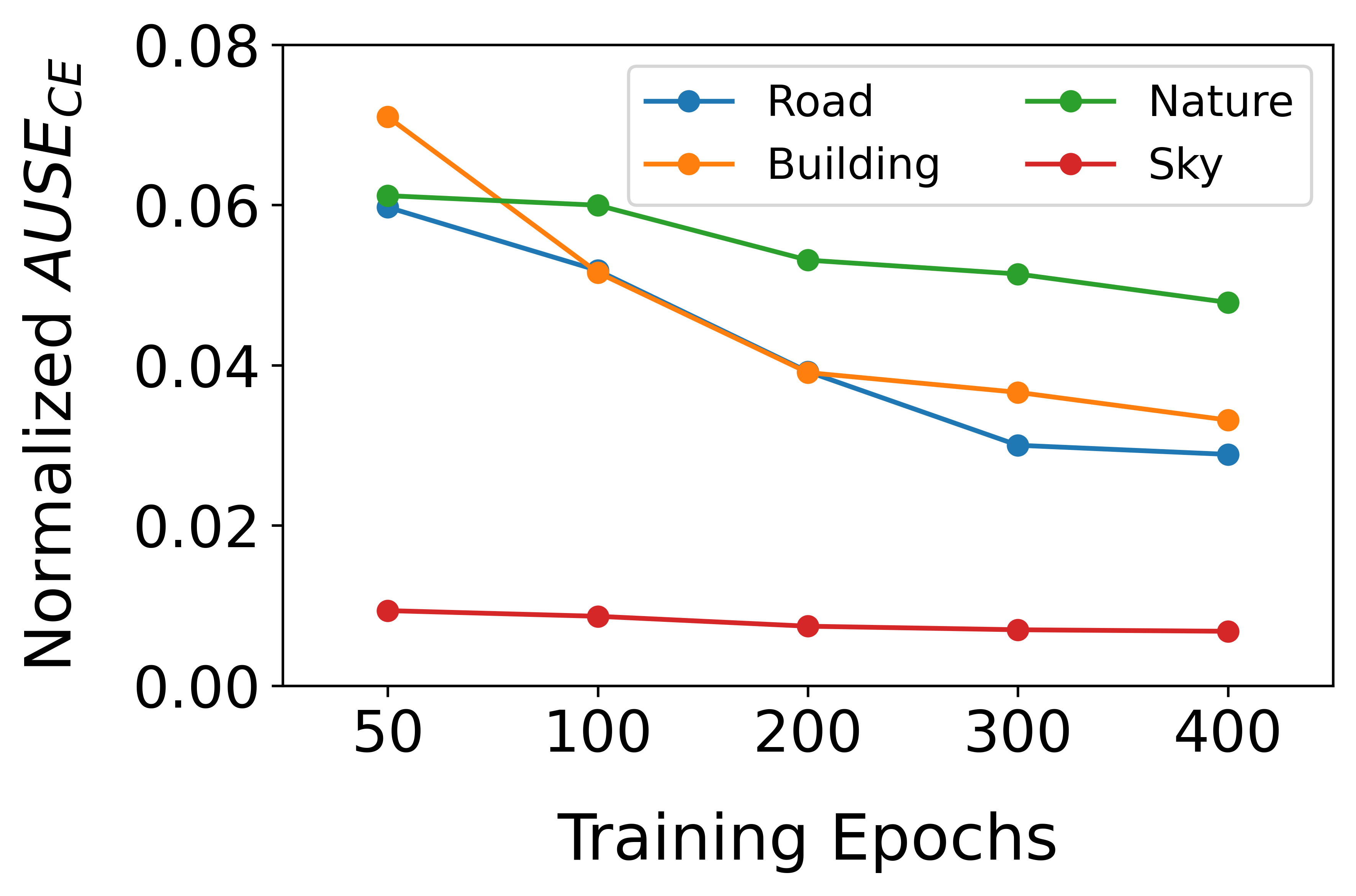}
    \end{minipage}
\vspace{-0.3cm}
\caption{Only classes with an occurrence of $>10^{7}$ are considered. It is evident that the class-wise AUSE$_{CE}$ converges to the lowest value for the 400-epoch model.}
\vspace{-0.2cm}
\label{fig:ause_ce}
\end{figure}

Typically, the cross-entropy is used as a suitable loss function for multi-class semantic segmentation. If an ideal uncertainty measure is available, then the sparsification curve is expected to align with the oracle curve suppressing the AUSE to zero. Hence, we propose the AUSE, obtained when cross-entropy is employed as the uncertainty measure to sort the predictions, as an estimator for the residual uncertainty. The residual uncertainty is non-reducible if the network architecture is fixed and comprises the aleatoric uncertainty of the underlying data generation process as well as the model uncertainty due to the limitations in the hypothesis space induced by the choice of the neural network architecture. This holds true if the model training converges. It is to be expected that the epistemic uncertainty, given by the sum of the approximation uncertainty and the model uncertainty~\cite{Uncertainty_decomposition}, reduces as model training improves until overfitting begins to occur. To support this hypothesis, we trained five models with an increasing number of epochs maintaining equivalence w.r.t. the model architecture and the hyperparameters respectively. The most prevalent classes, based on their representation in the chosen evaluation set, are considered for the evaluation of the AUSE utilizing cross-entropy~(AUSE$_{CE}$). The AUSE over the number of epochs is plotted in Figure~\ref{fig:ause_ce} and indicates strictly monotonic decreasing graphs, which supports the hypothesis.

\section{Conclusion}
With our work, we show evidence on the inconsistency between well-established calibration measures, which ultimately prevents the homologation of safety-critical applications based on neural networks. In detail, we studied the dependency of the optimal temperature in temperature scaling for a UNET model on the A2D2 dataset. We found that the optimal temperature retrieved by minimizing the NLL does not match the minimum of the ECE, as observed previously by Nixon et al.~\cite{nixon2019measuring}. Furthermore, we noticed a similar decoupling between the AUSE and the ECE, indicating a degree of freedom in calibrating neural networks. This undermines the goal of determining unique prediction intervals. We also empirically demonstrate that there is no difference in minimizing the ECE or UCE. Both methods lead to the same optimal temperature, which is remarkable considering that the UCE uses the entire prediction vector and the ECE relies exclusively on the probability mass allocated to the most likely class. In contrast, for the AUSE as a calibration metric, we also noticed a decoupling between the AUSE$_{V}$, using the point-wise uncertainty quantified by the variation ratio, and the AUSE$_{S}$, utilizing the Shannon entropy based on the entire prediction vector. As a result, current calibration metrics lead to different prediction intervals, generating an ambiguity that compromises trustworthiness. This conclusion also holds true if the ECE is compared to the CCQS or if their uncertainty-based counterparts, the UCE and UCQS are contrasted. For this, we transform the idea of assessing the areal deviation in the predicted confidences versus expected confidences graph in a regression setting to the reliability diagram in a classification problem. The ambiguity between these calibration measures arises from the skewness of the frequency distribution in the predicted confidences/uncertainties.

In addition, we propose the AUSE$_{CE}$ as a measure of the residual uncertainty if predictions are sparsified according to the cross-entropy. If model-training has converged to the best possible minimum within the loss landscape, the residual uncertainty quantifies the remaining uncertainty originating from the data generation process~(aleatoric nature) as well as the model uncertainty due to limitations in the hypothesis space given a fixed neural network architecture~(epistemic nature). Consequently, the model bias can be isolated by subtracting the conditional entropy, which makes the model uncertainty accessible.

\newpage
\bibliographystyle{splncs04}
\bibliography{library}

\begin{thebibliography}{10}
\providecommand{\url}[1]{\texttt{#1}}
\providecommand{\urlprefix}{URL }
\providecommand{\doi}[1]{https://doi.org/#1}

\bibitem{bishop2006pattern}
Bishop, C.M., Nasrabadi, N.M.: Pattern recognition and machine learning, vol.~4. Springer (2006)

\bibitem{Cordts2016Cityscapes}
Cordts, M., Omran, M., Ramos, S., Rehfeld, T., Enzweiler, M., Benenson, R., Franke, U., Roth, S., Schiele, B.: The cityscapes dataset for semantic urban scene understanding. In: Proc. of the IEEE Conference on Computer Vision and Pattern Recognition (CVPR) (2016)

\bibitem{cover1999elements}
Cover, T.M.: Elements of information theory. John Wiley \& Sons (1999)

\bibitem{Laplace_approximation}
Daxberger, E., Kristiadi, A., Immer, A., Eschenhagen, R., Bauer, M., Hennig, P.: Laplace redux - effortless bayesian deep learning  \textbf{34},  20089--20103 (2021), \url{https://proceedings.neurips.cc/paper_files/paper/2021/file/a7c9585703d275249f30a088cebba0ad-Paper.pdf}

\bibitem{Depeweg2017DecompositionOU}
Depeweg, S., Hern{\'a}ndez-Lobato, J.M., Doshi-Velez, F., Udluft, S.: Decomposition of uncertainty in bayesian deep learning for efficient and risk-sensitive learning. In: International Conference on Machine Learning (2017), \url{https://api.semanticscholar.org/CorpusID:3461501}

\bibitem{dreissig2022calibration}
Dreissig, M., Piewak, F., Boedecker, J.: On the calibration of underrepresented classes in lidar-based semantic segmentation (2022)

\bibitem{dreissig2023calibration}
Dreissig, M., Piewak, F., Boedecker, J.: On the calibration of uncertainty estimation in lidar-based semantic segmentation. In: IEEE 26th International Conference on Intelligent Transportation Systems (ITSC). pp. 4798--4805. IEEE (2023)

\bibitem{A2D2}
Geyer, J., Kassahun, Y., Mahmudi, M., Ricou, X., Durgesh, R., Chung, A.S., Hauswald, L., Pham, V.H., M{\"u}hlegg, M., Dorn, S., Fernandez, T., J{\"a}nicke, M., Mirashi, S., Savani, C., Sturm, M., Vorobiov, O., Oelker, M., Garreis, S., Schuberth, P.: {A2D2: Audi Autonomous Driving Dataset}  (2020), \url{https://www.a2d2.audi}

\bibitem{gneiting2007strictly}
Gneiting, T., Raftery, A.E.: Strictly proper scoring rules, prediction, and estimation. Journal of the American statistical Association  \textbf{102}(477),  359--378 (2007)

\bibitem{guo2017calibration}
Guo, C., Pleiss, G., Sun, Y., Weinberger, K.Q.: On calibration of modern neural networks pp. 1321--1330 (2017)

\bibitem{Gustafsson_2020_CVPR_Workshops}
Gustafsson, F.K., Danelljan, M., Schon, T.B.: Evaluating scalable bayesian deep learning methods for robust computer vision. In: Proceedings of the IEEE/CVF Conference on Computer Vision and Pattern Recognition (CVPR) Workshops (June 2020)

\bibitem{Uncertainty_decomposition}
Huellermeier, E., Waegeman, W.: Aleatoric and epistemic uncertainty in machine learning: an introduction to concepts and methods. Machine Learning  \textbf{110} (2021), \url{https://doi.org/10.1007/s10994-021-05946-3}

\bibitem{KOR}
Kim, T., Yun, S.Y.: Revisiting orthogonality regularization: A study for convolutional neural networks in image classification. IEEE Access  \textbf{10},  69741--69749 (2022). \doi{10.1109/ACCESS.2022.3185621}

\bibitem{kull2019beyond}
Kull, M., Perello~Nieto, M., K{\"a}ngsepp, M., Silva~Filho, T., Song, H., Flach, P.: Beyond temperature scaling: Obtaining well-calibrated multi-class probabilities with dirichlet calibration. Advances in neural information processing systems  \textbf{32} (2019)

\bibitem{Deep_Ensembles}
Lakshminarayanan, B., Pritzel, A., Blundell, C.: Simple and scalable predictive uncertainty estimation using deep ensembles  \textbf{30} (2017), \url{https://proceedings.neurips.cc/paper_files/paper/2017/file/9ef2ed4b7fd2c810847ffa5fa85bce38-Paper.pdf}

\bibitem{laves2019well}
Laves, M.H., Ihler, S., Kortmann, K.P., Ortmaier, T.: Well-calibrated model uncertainty with temperature scaling for dropout variational inference. arXiv preprint arXiv:1909.13550  (2019)

\bibitem{maag2020time}
Maag, K., Rottmann, M., Gottschalk, H.: Time-dynamic estimates of the reliability of deep semantic segmentation networks pp. 502--509 (2020)

\bibitem{Uncertainty_Disentanglement}
Mucsanyi, B., Kirchhof, M., Oh, S.J.: Benchmarking uncertainty disentanglement: Specialized uncertainties for specialized tasks  (2024), \url{https://arxiv.org/abs/2402.19460}

\bibitem{mukhoti2020calibrating}
Mukhoti, J., Kulharia, V., Sanyal, A., Golodetz, S., Torr, P., Dokania, P.: Calibrating deep neural networks using focal loss. Advances in Neural Information Processing Systems  \textbf{33},  15288--15299 (2020)

\bibitem{muller2019does}
M{\"u}ller, R., Kornblith, S., Hinton, G.E.: When does label smoothing help? Advances in neural information processing systems  \textbf{32} (2019)

\bibitem{nixon2019measuring}
Nixon, J., Dusenberry, M.W., Zhang, L., Jerfel, G., Tran, D.: Measuring calibration in deep learning. In: CVPR workshops. vol.~2 (2019)

\bibitem{mECE_mUCE}
Pakdaman~Naeini, M., Cooper, G., Hauskrecht, M.: Obtaining well calibrated probabilities using bayesian binning. Proceedings of the AAAI Conference on Artificial Intelligence  \textbf{29}(1) (Feb 2015). \doi{10.1609/aaai.v29i1.9602}, \url{https://ojs.aaai.org/index.php/AAAI/article/view/9602}

\bibitem{platt1999probabilistic}
Platt, J., et~al.: Probabilistic outputs for support vector machines and comparisons to regularized likelihood methods. Advances in large margin classifiers  \textbf{10}(3),  61--74 (1999)

\bibitem{UNET}
Ronneberger, O., Fischer, P., Brox, T.: U-net: Convolutional networks for biomedical image segmentation pp. 234--241 (2015)

\bibitem{Dropout}
Srivastava, N., Hinton, G., Krizhevsky, A., Sutskever, I., Salakhutdinov, R.: Dropout: A simple way to prevent neural networks from overfitting. Journal of Machine Learning Research  \textbf{15}(56),  1929--1958 (2014), \url{http://jmlr.org/papers/v15/srivastava14a.html}

\bibitem{tomani2022parameterized}
Tomani, C., Cremers, D., Buettner, F.: Parameterized temperature scaling for boosting the expressive power in post-hoc uncertainty calibration. In: European Conference on Computer Vision. pp. 555--569. Springer (2022)

\bibitem{Wolf_ICCV2023}
Wolf, D.W., Ulrich, M., Kapoor, N.: Sensitivity analysis of ai-based algorithms for autonomous driving on optical wavefront aberrations induced by the windshield. 2023 IEEE/CVF International Conference on Computer Vision Workshops (ICCVW) pp. 4102--4111 (10 2023). \doi{10.1109/ICCVW60793.2023.00443}, \url{https://ieeexplore.ieee.org/document/10350923/}

\bibitem{wursthorn2024uncertainty}
Wursthorn, K., Hillemann, M., Ulrich, M.: Uncertainty quantification with deep ensembles for 6d object pose estimation. arXiv preprint arXiv:2403.07741  (2024)

\end{thebibliography}
\end{document}